# A MULTI-WEIGHT SELF-MATCHING VISUAL EXPLANATION FOR CNNS ON SAR IMAGES

*Siyuan Sun[1*], Yongping Zhang[2], Hongcheng Zeng[1], Yamin Wang[1], Wei Yang[1], Wanting Yang[1], Jie Chen[1]*

[1]*The School of Electronic and Information, Engineering, Beihang University, Beijing, China*
[2]*Mianyang Tianyi Space Technology Co., Ltd, Sichuan China*
**ssy2001@buaa.edu.cn*

**Keywords**: Synthetic Aperture Radar(SAR), Convolutional Neural Networks(CNNs), Class Activation Mapping, Visual Explanation, Weakly-supervised Object Localization

## Abstract

In recent years, convolutional neural networks (CNNs) have achieved significant success in various synthetic aperture radar (SAR) tasks. However, the complexity and opacity of their internal mechanisms hinder the fulfillment of high-reliability requirements, thereby limiting their application in SAR. Improving the interpretability of CNNs is thus of great importance for their development and deployment in SAR. In this paper, a visual explanation method termed multi-weight self-matching class activation mapping (MS-CAM) is proposed. MS-CAM matches SAR images with the feature maps and corresponding gradients extracted by the CNN, and combines both channel-wise and element-wise weights to visualize the decision basis learned by the model in SAR images. Extensive experiments conducted on a self-constructed SAR target classification dataset demonstrate that MS-CAM more accurately highlights the network's regions of interest and captures detailed target feature information, thereby enhancing network interpretability. Furthermore, the feasibility of applying MS-CAM to weakly-supervised object localization is validated. Key factors affecting localization accuracy, such as pixel thresholds, are analyzed in depth to inform future work.

## 1 Introduction

Synthetic aperture radar (SAR) offers all-weather, all-day, wide-area, and high-resolution imaging capabilities, and has been widely applied in surface monitoring, military reconnaissance, urban planning, and disaster assessment [1-3]. With the continuous advancement of imaging technologies, SAR imagery is increasingly characterized by massive volumes and high precision. In contrast, the interpretation capability for SAR images remains lagging, particularly in high-precision target recognition tasks, which still face significant challenges. Traditional SAR target recognition relies on a multi-stage image processing pipeline-including denoising, edge detection, feature extraction, and handcrafted classification rules. Such a pipeline is inefficient, computationally burdensome, and limited in adaptability and scalability, making it difficult to satisfy real-time requirements in practical SAR applications.

Deep learning represented by deep neural networks has achieved remarkable progress in recent years. As one of the representative models of deep learning, convolutional neural networks (CNNs) have obtained outstanding results in computer vision and have been introduced into SAR image interpretation tasks. CNNs can automatically learn multi-level, abstract discriminative features from data without cumbersome preprocessing steps, and their end-to-end frameworks effectively improve recognition accuracy and processing efficiency in SAR target recognition tasks. However, due to their complex and opaque internal mechanisms, CNNs are often regarded as "black box" models, and their decision bases cannot be directly interpreted. This opacity poses application barriers in remote sensing, where high safety and reliability are required. Enhancing the interpretability of CNNs in SAR image processing is therefore of paramount importance.

Class activation mapping (CAM)-based visual explanation methods have been proposed to address this issue. Such methods typically use feature maps from deep CNN layers and corresponding importance weights for specified classes to generate saliency maps that visualize the image regions on which the model focuses, thereby revealing the feature information underlying network decisions and improving interpretability. Zhou et al. introduced CAM, which visualizes model attention regions using the last convolutional layer's feature maps and a global average pooling layer [4]. Selvaraju et al. proposed gradient-weighted class activation mapping (Grad-CAM), which weights feature maps by gradient information to generate saliency maps [5].

Chattopadhay et al. optimized Grad-CAM and proposed Grad-CAM++, introducing higher-order derivatives to obtain richer information and achieve stronger boundary localization [6]. Wang et al. developed Score-CAM, which eschews gradient information and computes feature map importance weights via multiple forward passes [7]; this improves visual explanation quality at the expense of increased computational load. Jiang et al. applied element-wise weights to feature maps, better accounting for the contribution of each spatial location [8].

However, SAR images possess unique imaging mechanisms, and CAM methods designed for optical imagery often yield suboptimal visual explanations on SAR data, failing to accurately capture targets' spatial energy distributions and structural features. In this context, Feng et al. proposed Self-Matching CAM, which matches the input SAR image with selected feature maps to generate high-quality visual explanations [9]; subsequent work optimized computational efficiency [10, 11]. Nevertheless, the resulting saliency maps still leave room for improvement in terms of attention region completeness and target detail preservation.

In this paper, a multi-weight self-matching class activation mapping (MS-CAM) method tailored for SAR imagery is proposed to address these limitations. MS-CAM builds on self-matching between the input image and intermediate feature maps, and introduces a joint weighting mechanism at both element-wise and channel-wise levels to generate more precise saliency maps. These saliency maps more clearly depict the regions and fine-grained features attended by the CNN during decision making. To validate the effectiveness of the proposed method, a SAR target classification dataset was constructed and employed. Experimental results demonstrate that MS-CAM more completely and accurately reflects the key attention regions in SAR images during CNN decision making, while better preserving target detail. Finally, the feasibility of weakly supervised object localization based on MS-CAM is verified, and the localization results are presented and analyzed.

## 2 Methodology

The multi-weight self-matching class activation mapping (MS-CAM) learns the Self-Matching CAM concept, introduces a self-matching mechanism, and simultaneously applies element-wise and channel-wise weights to feature maps to generate more refined saliency maps, as detailed below.

For a set of $G\times G$ feature maps $A$ output by a selected network layer, element-wise weights are first computed. The gradient $\partial y^c / \partial A^k$ of the *k-th* feature map $A^k$ with respect to the output $y^c$ of the final fully connected layer for class $c$ is obtained. The gradient is then passed through a ReLU activation function, retaining positive values and discarding negative values, thereby yielding the element-wise weight $\omega_{ele}^{kc}$, as follows:

$$\omega_{ele}^{kc} = \mathrm{ReLU}(\frac{\partial y^c}{\partial A_G^k}) \tag{1}$$

where $\omega_{ele}^{kc}$ denotes the element-wise weight of the *k-th* feature map $A^k$, also of size $G\times G$; $c$ denotes the specified class, and $\mathrm{ReLU}(\cdot)$ is the activation function. Thereafter, a Hadamard product is applied between each feature map and its corresponding element-wise weight to produce the weighted feature maps, as shown below:

$$\hat{A}_G^k = w_{ele}^{kc} \circ A_G^k \tag{2}$$

where $\hat{A}_G^k$ denotes the weighted feature map obtained; the number and dimensions of the feature maps remain unchanged after weighting. Element-wise weights account for the importance of each spatial position within a single feature map, and the processing in Equation (2) contributes to reducing and alleviating background noise in the saliency maps.

After $\hat{A}_G^k$ is obtained, the input image $I$ of size $N\times N$ is downsampled to $M\times M$, and $\hat{A}_G^k$ is upsampled to $M\times M$, where the intermediate size $M$ satisfies $G < N$ and $M \in [G, N]$. Both downsampled and upsampled maps are then normalized. The chosen intermediate size avoids introducing excessive irrelevant information during upsampling while preserving more original image details, as expressed by the following equation:

$$I_M = s(D(I_N)_M) \tag{3}$$

$$\hat{A}_M^k = s(U(\hat{A}_G^k)_M) \tag{4}$$

where $I_N$ denotes the input image, $D(\cdot)$ the downsampling function, $U(\cdot)$ the upsampling function, $s(\cdot)$ the normalization function, and $I_M$ and $\hat{A}_M^k$ denote, respectively, the input SAR image and the *k-th* feature map after the two processing steps. Subsequently, each processed feature map $\hat{A}_M^k$ is subjected to a Hadamard product with $I_M$ to fully exploit the feature information contained in the input SAR image, as follows:

$$\tilde{A}_M^k = I_M \circ \hat{A}_M^k \tag{5}$$

where $\tilde{A}_M^k$ denotes the processed new feature map. Subsequently, $\tilde{A}_M^k$ is upsampled to $N\times N$, weighted and summed across channels using channel-wise weights, passed through a ReLU activation, and finally normalized to yield the final saliency map $L_{MS-CAM}^c$, as shown below:

$$L_{MS-CAM}^c = s(\mathrm{ReLU}(\sum_k \omega_{cha}^{kc} Up(\tilde{A}_M^k)_N)) \tag{6}$$

where $\omega_{cha}^{kc}$ denotes the channel-wise weight of the feature maps, which can be obtained by any CAM method. In the case of Grad-CAM, the gradients are globally average pooled to derive the channel-wise weights $\omega_{cha}^{kc}$ :

$$\omega_{cha}^{kc} = \frac{1}{G^2}\sum_{i}\sum_{j}\frac{\partial y^c}{\partial A_{ij}^k} \tag{7}$$

## 3 Results

### *3.1 Experimental Setup*

A dataset for SAR target classification was constructed to validate the effectiveness of MS-CAM, and its basic attributes are summarized in Table 1.

Table 1 Dataset Details

| Source | Resolution | Mode | Band | Size |
|---|---|---|---|---|
| Gaofen-3 | 1~3m | Spotlight, Stripmap | C | 512×512 |

The dataset comprises four classes: container ship, bulk carrier, oil tanker, and aircraft, with examples of each shown in Figure 1. Each class contains 1,000 SAR images of size 512×512 pixels. The dataset was partitioned into training and testing sets at a ratio of 4:1, yielding 800 images per class for training and 200 for testing. Training was performed on ResNet50; cross-entropy loss was employed, the initial learning rate was set to $5 \times 10^{-1}$, the momentum coefficient was set to 0.9, and stochastic gradient descent was used as the optimizer. The number of training epochs was set to 50.

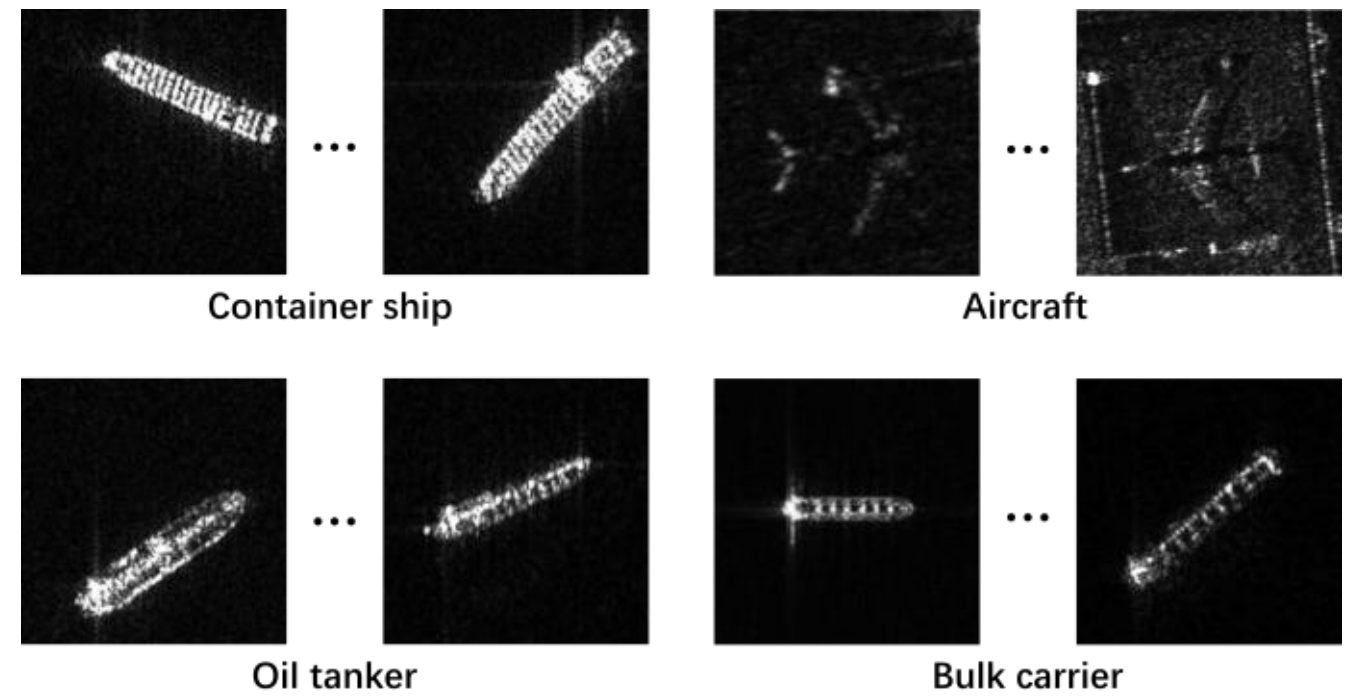

Figure 1 Examples of data

### *3.2 Analysis of Visual Explanation Results*

First, based on the trained ResNet50, saliency maps were generated using Grad-CAM, Grad-CAM++, Score-CAM, Layer-CAM, Self-Matching CAM, and MS-CAM for multiple SAR images to compare the visual explanation performance of these methods and validate the effectiveness of MS-CAM. Fourteen images covering four classes were selected for the experiment. The saliency maps produced by each method were converted to heatmaps for better visual comparison; results are presented in Figure 2.

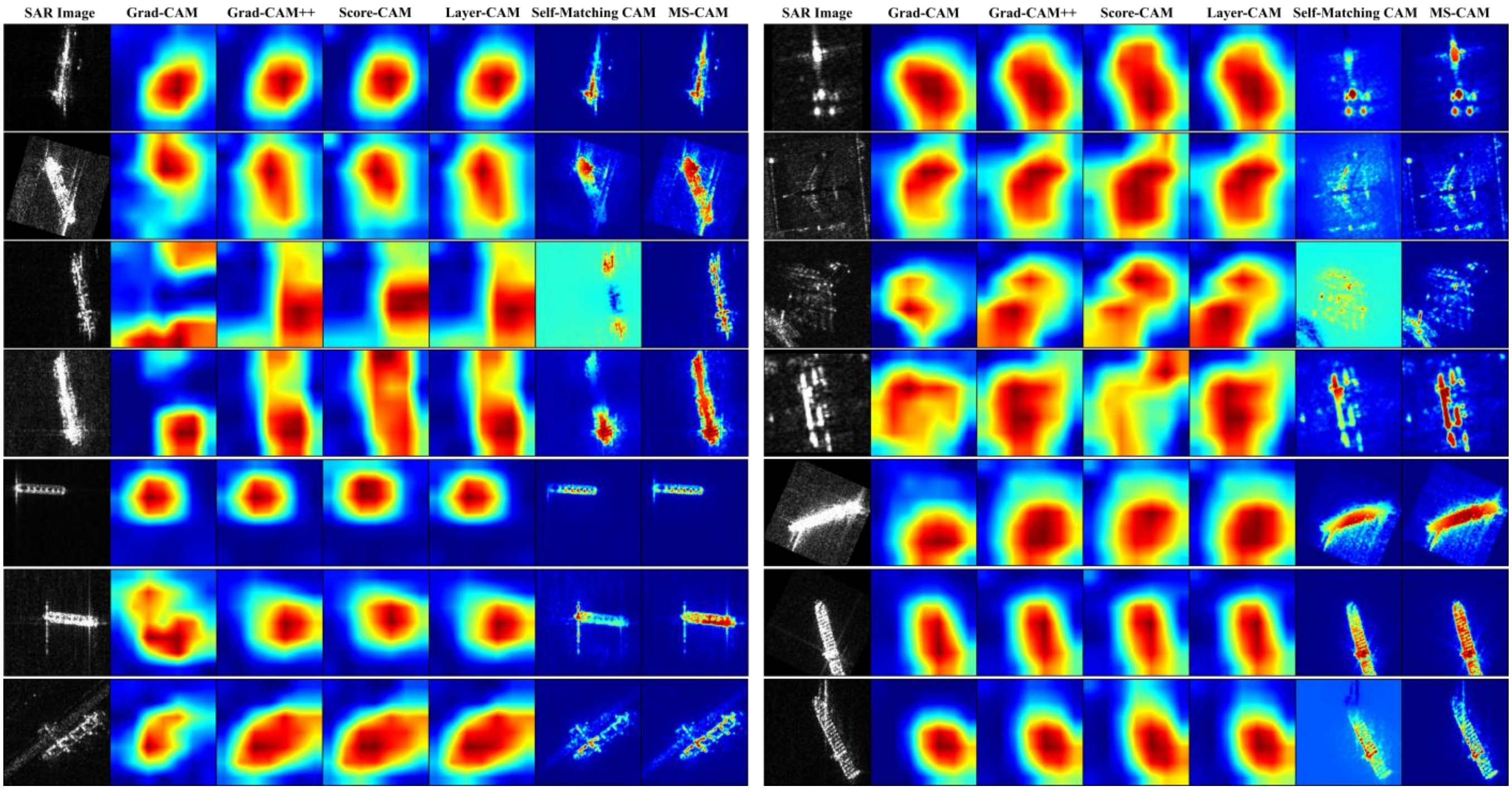

Figure 2 Comparison of visual explanation results

The results indicate that, except for Self-Matching CAM and MS-CAM, methods originally designed for optical imagery, such as Grad-CAM, yield over-diffused highlighted regions and struggle to cover the entire target. Grad-CAM even focuses on background areas in some images, likely due to the unique imaging mechanism of SAR. Both Self-Matching CAM and MS-CAM accurately localize target regions; however, the high-intensity red regions in Self-Matching CAM often cover only parts of the targets, whereas MS-CAM almost entirely covers the targets and better emphasizes fine details, such as the bow and stern of ships and pipelines on oil tankers. Additionally, Self-Matching CAM exhibits excessive background noise in certain cases, while MS-CAM reduces background noise intensity through element-wise weighting, thereby more clearly highlighting the SAR regions attended during model decision-making.

The visual explanation capability of MS-CAM at shallow model layers was analyzed using ResNet50. Shallow layers often attend to both target and background features, and relying solely on channel-wise weights fails to suppress background noise effectively, leading to saliency maps that cannot accurately localize targets. By combining element-wise and channel-wise weights, MS-CAM more faithfully reflects the importance of each channel and each spatial position in the feature maps for model decisions.

In the experiment, ResNet50 was divided into four hierarchical levels based on its architecture, and the features captured at each level were visualized using Layer-CAM and MS-CAM. Methods such as Self-Matching CAM, which employ only channel-wise weights, were excluded from this comparison due to their suboptimal performance at shallow layers. The results, shown in Figure 3, indicate that both Layer-CAM and MS-CAM can visualize shallow-layer features; however, MS-CAM produces saliency maps with lower noise intensity and more finely delineates the target regions on which the network focuses. Moreover, as the network depth increases, Layer-CAM's highlighted regions tend to diffuse, whereas MS-CAM remains concentrated on the target regions and better captures the detailed features learned by the model.

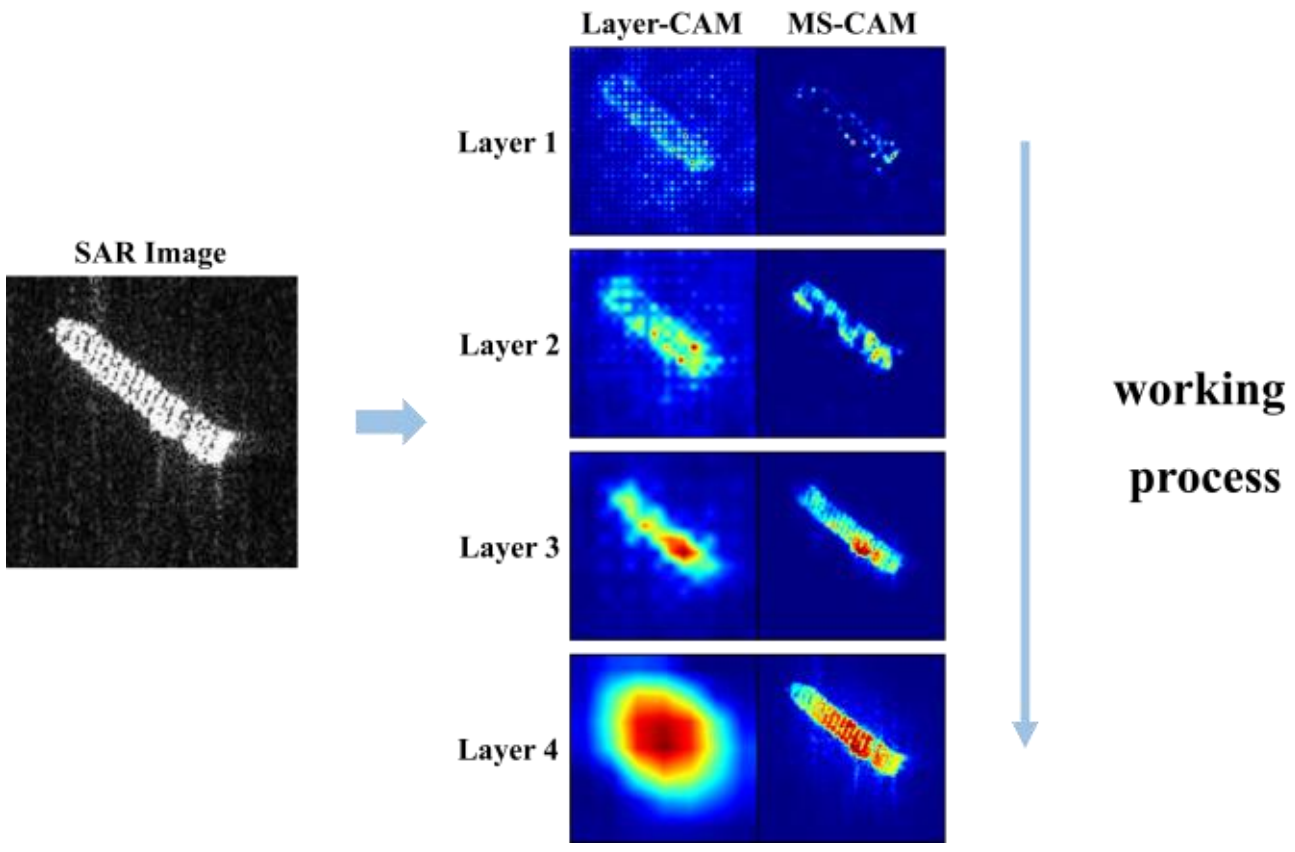

Figure 3 Visual explanation result of shallow-layer

### *3.3 Weakly-supervised Object Localization*

In this section, SAR image target localization is attempted using the classification model and MS-CAM, with the trained ResNet50 model employed. Upon input of a SAR image into the model, the predicted class is obtained. MS-CAM is then applied at the final layer of model to generate a saliency map for the predicted class. A pixel threshold is set to binarize the saliency map, producing connected pixel segments. A bounding box is drawn around the largest connected segment to achieve target localization. The processing workflow and final results are shown in Figure 4.

In the processing shown in Figure 4, a threshold set to 45% of the maximum pixel value yielded satisfactory localization results. Varying this threshold produced the results in Figure 5, demonstrating that threshold selection affects localization accuracy. When the threshold was reduced to 30% of the maximum pixel value, the bounding box failed to align with the object boundary and included excessive background. Conversely, increasing the threshold caused the bounding box to cover only part of the object.

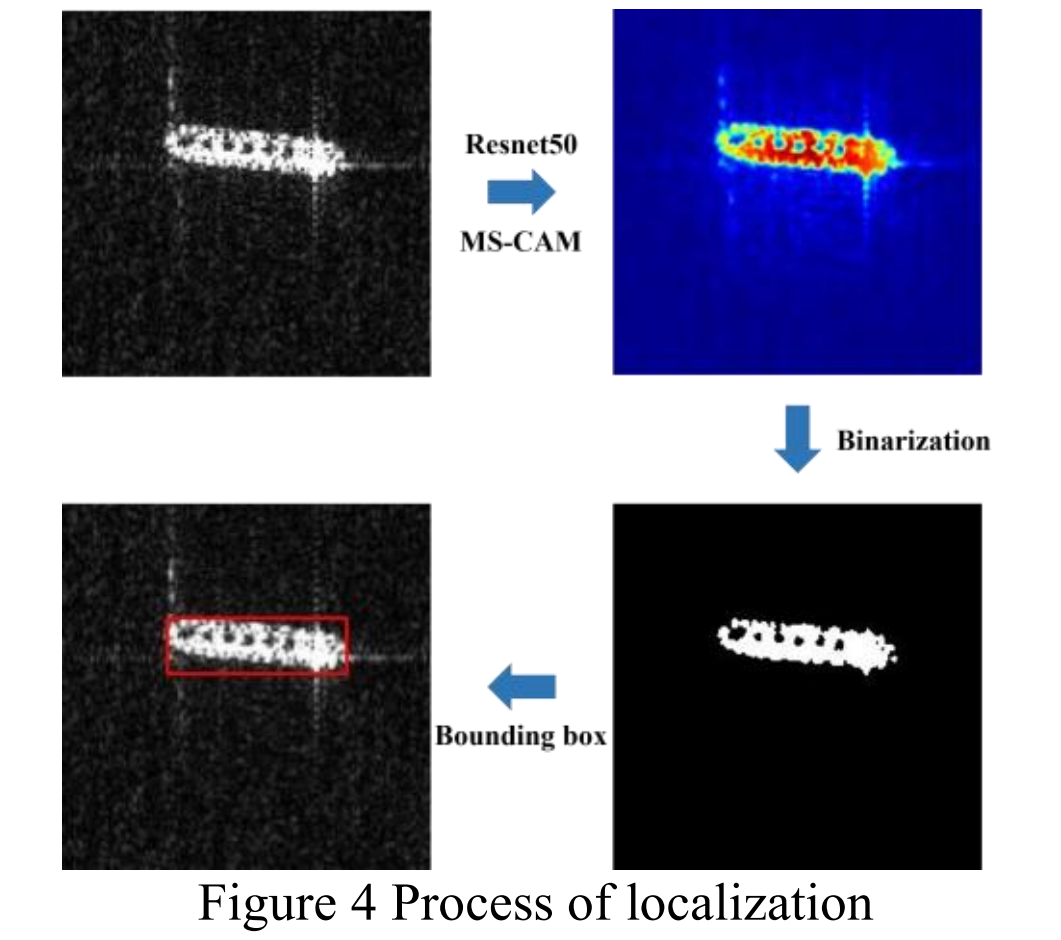

Figure 4 Process of localization

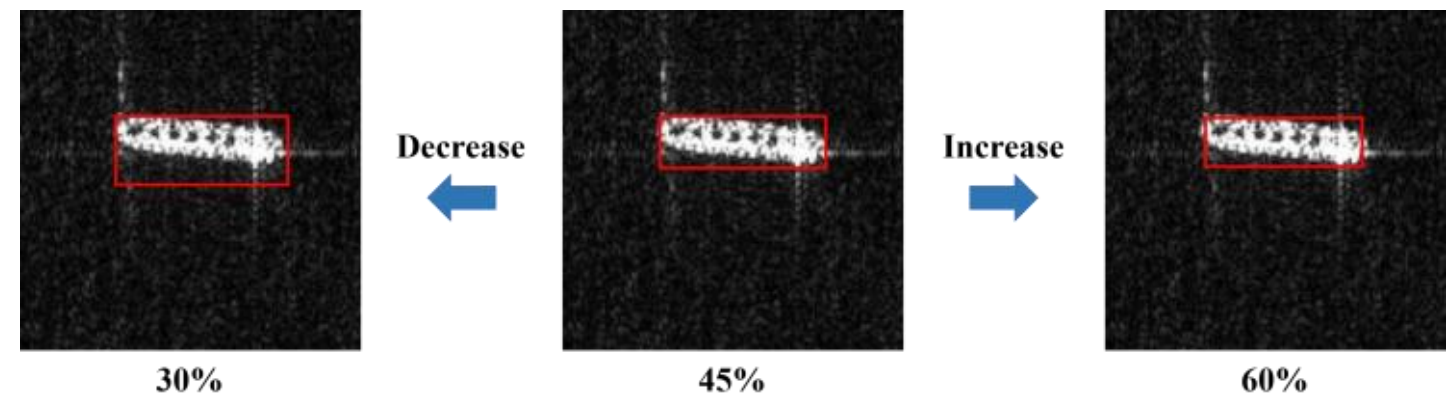

Figure 5 Localization results at different thresholds

Additionally, Weakly supervised object localization was performed on various images to assess the consistency of threshold settings. The results shown in Figure 6 highlight, with blue arrows, the optimal localization outcome for each image under three threshold values. Significant variation in optimal thresholds among images is evident. Consequently, employing a fixed threshold for batch processing may result in poor localization performance for some images. Dynamic threshold adjustment may be considered in future work to better accommodate the varying threshold requirements of different images and thus improve overall localization performance.

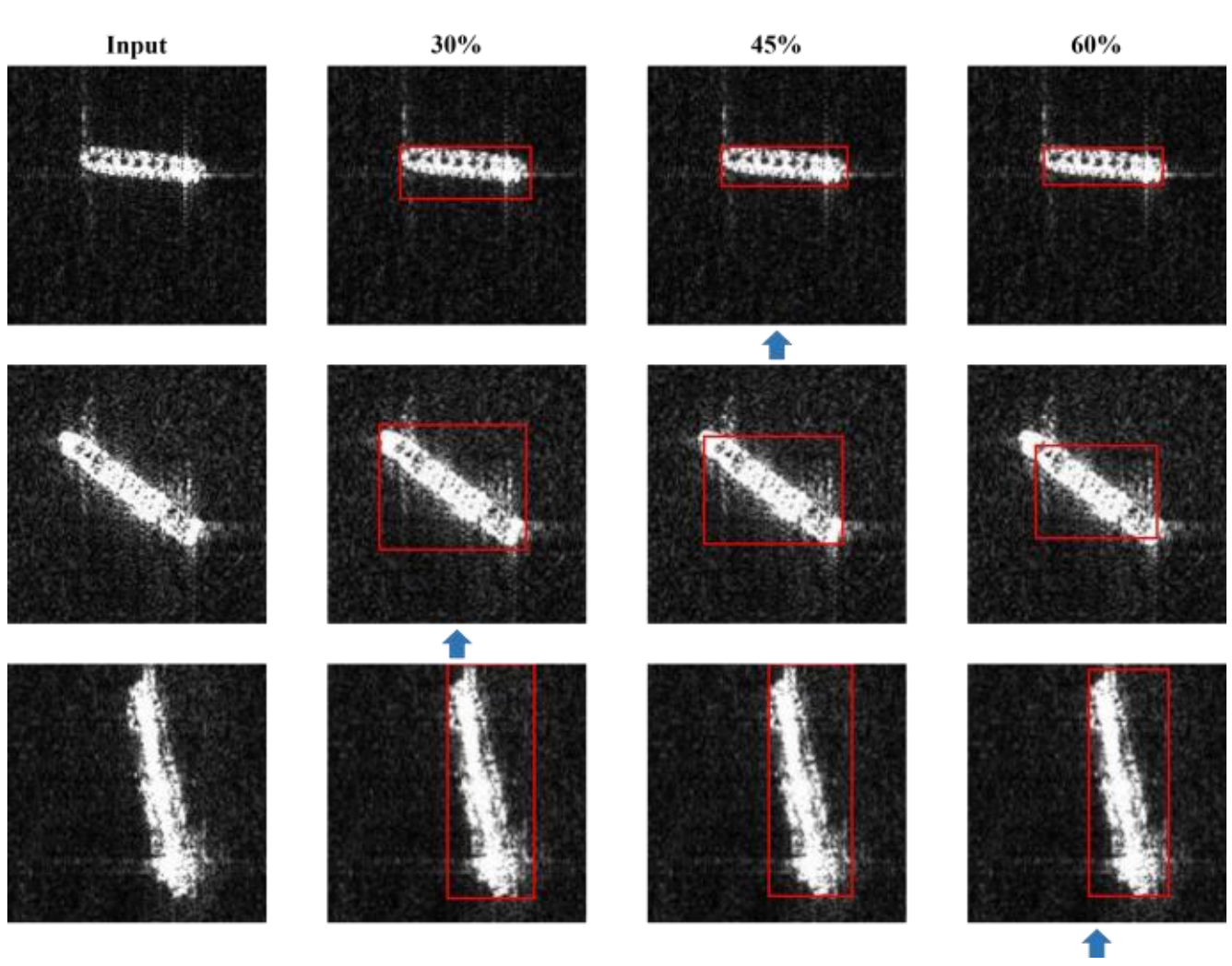

Figure 6 Localization results of different images at three thresholds (Arrows Indicate the Best Results)

## 4 Conclusion

In this paper, a visual explanation method MS-CAM is proposed for CNNs applied to various SAR image tasks. MS-CAM is implemented based on class activation mapping. Compared to other CAM methods designed for optical imagery, MS-CAM is able to more accurately focus on target regions in SAR images; compared to the Self-Matching CAM method, which performs well on SAR images, MS-CAM is capable of reflecting more complete network attention regions as well as finer target details. Additionally, relative to the aforementioned methods, MS-CAM exhibits superior visual explanation capability in the shallow layers of model, better illustrating the process by which the model extracts target feature information from SAR images and enhancing the interpretability of network. Weakly supervised object localization was also achieved based on MS-CAM and the classification model, with analysis of the maximum impact of pixel threshold on localization accuracy. Future work may consider dynamic threshold adjustment to improve overall localization performance.

## 5 Acknowledgements

This work is supported by the National Science Foundation of China (NFSC) under Grant U23B2007.